\documentclass{article}

\usepackage{times}
\usepackage{graphicx} 
\usepackage{subfigure} 
\usepackage{amsmath}
\usepackage{amssymb}
\usepackage{natbib}
\usepackage{amsthm}
\usepackage{comment}
\usepackage{array}
\newcolumntype{P}[1]{>{\centering\arraybackslash}p{#1}}

\usepackage{algorithm}
\usepackage{algorithmic}

\usepackage{hyperref}

\usepackage[accepted]{icml2016}

\icmltitlerunning{Image Generation and Editing with Variational Info Generative Adversarial Networks}

\begin{document} 

\twocolumn[
\icmltitle{Image Generation and Editing with Variational Info Generative Adversarial Networks}

\icmlauthor{Mahesh Gorijala}{maheshgorijala@ee.iisc.ernet.in}
\icmlauthor{Ambedkar Dukkipati}{ad@csa.iisc.ernet.in}
\icmladdress{Department of Computer Science and Automation \\
Indian Institute of Science, Bangalore-560012, India}

\icmlkeywords{boring formatting information, machine learning, ICML}

\vskip 0.3in
]
\begin{abstract}
Recently there has been an enormous interest in generative models for images in deep  learning. In pursuit of this, Generative Adversarial Networks (GAN) and Variational Auto-Encoder (VAE) have surfaced as two most prominent and popular models.  While VAEs tend to produce excellent reconstructions but blurry samples, GANs generate sharp but slightly distorted images. In this paper we propose a new model called Variational InfoGAN (ViGAN). Our aim is two fold: (i) To generated new images conditioned on visual descriptions, and (ii) modify the image, by fixing the latent representation of image and varying the visual description. We evaluate our model on Labeled Faces in the Wild (LFW), celebA and a modified version of MNIST  datasets and demonstrate the ability of our model to generate new images as well as to modify a given image by changing attributes.   
\end{abstract} 

\section{Introduction}
Generative models, that are essential part of unsupervised learning, capture the structure/patterns in data by learning to generate samples that resemble the training data. 
Earlier work in generative modeling focused on graphical models  and energy based
models with latent variables, for example, Restricted Boltzmann Machines (RBM)~\cite{bengio2011unsupervised, hinton2010practical}, Deep Belief Networks (DBN)~\cite{hinton2006fast}. In these models, exact inference, normalization constant and its gradients are intractable and  samples are usually obtained with expensive Markov Chain Monte  Carlo (MCMC) techniques. 
   
Recent advances in deep learning have enabled us to bypass above mentioned challenges by using  deep neural networks as parametrized functions that generate samples. Few dominant approaches that emerged in recent years are Variational AutoEncoders (VAEs)~\cite{AEVB}, Generative Adversarial Networks (GANs)~\cite{GAN}, Generative Stochastic Networks~\cite{GSN}, Deep Recurrent Attention Writer~\cite{DRAW}, Pixel Recurrent Neural Networks~\cite{pixelRNN} and Pixel Convolutional Neural Networks~\cite{pixelCNN}. Generative models have also helped to set benchmark results in semi-supervised learning  \cite{CVAE, DCGAN}. Lately there have been attempts to use the advances in deep generative models for semi-supervised learning~\cite{CVAE,InfoGAN} and joint modeling of images and their visual descriptions~\cite{cGAN, aGAN}. 
   
In this work, we address the problem of jointly modeling images and their visual descriptions through  deep generative models. Multimodal learning requires learning a correspondence between data across  different modalities. In deep learning context, the first works towards multimodal learning is by N. Srivatsava \textit{et al.}~\cite{sriRBM} and J. Nigam~\textit{et al.}\cite{ngMM}. These models  pose this problem as learning cross-modal representations at latent feature level and some models even try to learn a shared representation across modalities.
This often becomes complex, because one modality rarely contains all the information about the other modality. In the case of images and text, the information present in image modality is much more dense than the information in text modality.

VAEs are trained by maximizing the lower bound to the log-likelihood of the data and they include an inference network as part of the training procedure. Extensions such as~\cite{burda2015importance} are proposed to increase the tightness of the lower bound. On the other hand, GANs are trained as a mini-max game between the generator and discriminator. Training of GANs only require a differentiable mapping from a latent space to the data space, without any requirement for inference network.  
Recently proposed InfoGAN~\cite{InfoGAN} framework, a variation of GAN~\cite{GAN}, defines image generation as a function of two sets of variables and is shown to learn interpretable representations of images. This is achieved by including a mutual information objective between a small subset  of interpretable latent variables (denoted by $c$) and the observations. But as in GANs, it also  lacks an inference mechanism to infer the latent state given an image. 
   
In this paper, we propose a new model, Variational Info GAN (ViGAN) that combines the InfoGAN model with VAE by using a subset of latent variables to represent the description of images. We also extend the architecture to include an inference mechanism and present a training  procedure to jointly train the inference network and the model. We provide a training and sampling procedure for the proposed model. We denonstrate ViGAN's ability to generate new images conditioned on attributes, as well as modifying given images by changing attributes  on three datasets namely, MNIST~\cite{lecun1998mnist}, LFW~\cite{LFWTech}, celebA~\cite{celebA}. 
   
\section{Background Material}
VAEs and GANs are based on the idea of using a differentiable neural network to transform latent variables to data samples. While the training procedure for VAE includes an approximate inference network, GAN is trained via an auxiliary discriminator network, that learns to discriminate real data samples from generated samples. In this section, we briefly describe previous work on VAE, GAN and InfoGAN on which our model builds upon. In Section 3, we develop our proposed model using the notations introduced in this section. 

\subsection{Variational Auto-encoder (VAE)}
A VAE~\cite{AEVB} works on the principle of maximizing lower bound to the marginal likelihood of the data. In simple terms, VAE can be viewed as an autoencoder with a prior over 
latent variables. It consists of two networks, encoder, a mapping from data space $X$ to latent space $Z$ and decoder, a reverse mapping from latent space $Z$ to data space $X$ . 

Let $p(x|z)$  represent the probability distribution parametrized by the decoder, $p(z)$ represent the prior over the latent space and $p(x)$ represent probability of the data point $x$. Then upper bound on the negative log-likelihood of a data point $x$ can be written as 
\begin{align}
	-\log  p(x) &\leq  -\mathsf{E} _{q(z|x)} \log p(x|z) +  \mathrm{KL}(q(z|x) || p(z))  \nonumber \\
	 &\leq L_{\mathrm{recon}} + L_{\mathrm{prior}},
\end{align}
where $ q(z|x)$ is an approximation to the true posterior $ p(z|x)$.
The loss function for VAE is given as
\begin{equation}
	L_{\mathrm{VAE}} = L_{\mathrm{recon}} + L_{\mathrm{prior}},
\end{equation}
where $ L_{\mathrm{recon}} $ represent reconstruction loss and $ L_{\mathrm{prior}} $ represents KL divergence between the approximate posterior $ q(z|x) $ and the prior $ p(z) $. 

\subsection{Generative Adversarial Network (GAN)}
GAN \cite{GAN} consist of two networks, namely, generator ($G$) and discriminator ($D$). Generator network takes noise $z$ drawn from a prior distribution $ p(z)$ as input and give an image as output. Discriminator network gives the probability that the input image is real.

The GAN objective is to train discriminator to distinguish between real and fake data samples, while simultaneously training the generator to fool discriminator. Losses for the model are given as 
\begin{align}
  L_{\mathrm{gen}} &= \mathsf{E}_{z\sim p(z)}[-\log  D(G(z))],  \quad  
  \mbox{and} \\
	L_{\mathrm{dis}} &= \mathsf{E}_{z\sim p(z)}[\log  D(G(z))] + \mathsf{E}_{x\sim p(x)}[-\log  D(x)] ,
\end{align}
where $L_{\mathrm{gen}} $ and $L_{\mathrm{dis}} $ represent generator and discriminator losses respectively. During training, parameters of generator and discriminator are updated alternatively.  

InfoGAN \cite{InfoGAN} extends GAN by maximizing mutual information  between a small set of latent variables and observations. This can be incorporated into GAN objective by adding a loss term representing the mutual information. This ensures that the information represented by a set of ``interpretable  latent variables'' ($c$), is not lost during generation process. Let $ [z,c] $ be the input to the generator. Then mutual information between $c$ and $G(z, c) $ is
\begin{align}
	\mathsf{I}(c; G(z,c) &= \mathsf{H}(c) - \mathsf{H}(c|G(z,c)) \\ 
	 &\geq \mathsf{H}(c) + \mathsf{E}_{x \sim G(z,c)}[\mathsf{E}_{c \sim P(c|x)}[-\:\log \: Q(c|x)]],
\end{align}
where $ Q(c|x) $ is approximation to $P(c|x)$. This lower bound can be maximized directly. Hence recognition loss can be defined as 
\begin{align}
	L_{\mathrm{recog}} &= - \mathsf{E}_{x \sim G(z,c)}[\mathsf{E}_{c \sim P(c|x)}[-\:\log \: Q(c|x)]].
\end{align}

Now the modified GAN objective can be written as 
\begin{align}
	L_{\mathrm{gen}} &= E_{z \sim p(z)}[-\:\log \: D(G(z))] + L_{\mathrm{recog}}, \quad \mbox{and} \\ 
	L_{\mathrm{dis}} &= E_{z \sim p(z)}[\:\log \: D(G(z))] + \mathsf{E}_{x \sim p(x)}[-\:\log \: D(x)]. 
\end{align}
The recognizer $Q(c|x)$ is realized by a neural network. It can be made to share parameters with the discriminator. The set of interpretable latent variables $c$ can either be continuous, categorical or a combination of both. 
\section{Proposed Method}
\label{sec:pagestyle}
\subsection{Motivation}
Mutual information objective of InfoGAN ensures that the information represented by the interpretable latent variables ($c$), is preserved during the generation process. But it has no control over the information that $\mathbf{c}$ represents.   
In this paper, we propose a method to jointly model images and their descriptions by using $c$ to represent the known aspects of the data, and  $z$ to represent all the remaining factors of variation that are not captured by the attributes $c$. This can done by providing $(\tilde{z},c)$ as an input to the generator, where $\tilde{z}$ is the representation of image obtained by passing it through an encoder. 
In this setup, we assume that the description $c$ is available for all the images.
This approach differs from that of InfoGAN, where no labelled data is used and other semi-supervised approaches~\cite{CVAE}, in which only a subset of data is assumed to be labelled.  

\begin{figure*}
\begin{center}
  \centering
  \centerline{\includegraphics[height=9cm]{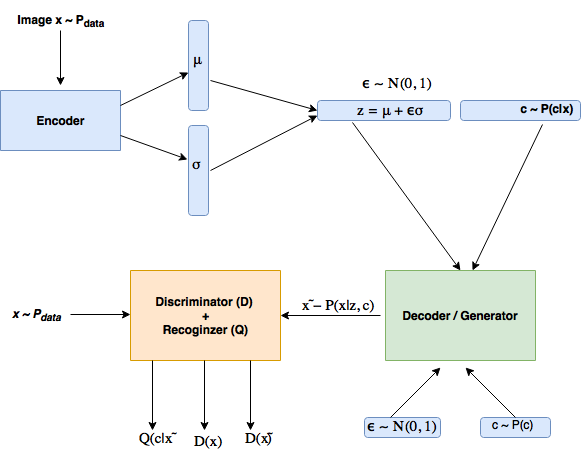}}
\end{center}
   \caption{Detailed schematic diagram of the proposed network. The proposed network
    combines the architectures of VAE and InfoGAN}
\label{fig:short}
\end{figure*}

The encoder-generator combination with a prior over $\tilde{z}$ can be thought of as an VAE, where as generator-discriminator combination can be taken to be GAN. Schematic representation of the network architecture is presented in Figure~\ref{fig:short}. Expectations over the approximate posterior $q(z|x)$ are computed using the reparametrization trick presented in~\cite{AEVB}.

\subsection{Training and Sampling}
The loss for the encoder network $ L_{\mathrm{enc}} $ is sum of reconstruction loss $ L_{\mathrm{recon}} $ and the KL divergence $L_{\mathrm{prior}}$ as described in the previous section:
\begin{align}
	L_{\mathrm{enc}} &= L_{\mathrm{prior}} + L_{\mathrm{recon}} ,  \\		
	L_{\mathrm{prior}} &= \mathrm{KL}(q(z|x) || p(z)) , \quad and \\ 
	L_{\mathrm{recon}} &=  -\mathsf{E}_{z \sim q(z|x) , c \sim P(c|x)} \log P(x|z,c).
\end{align}
Loss for the recognition network $Q$, denoted as $L_{\mathrm{recog}}$ is computed using only the samples   from the actual dataset.
\begin{align}
	L_{\mathrm{recog}} &= - \mathsf{E}_{x\sim P(x)}[\mathsf{E}_{c \sim P(c|x)}[-\:\log \: Q(c|x)]] 
\end{align}
Loss for the generator/decoder is comprised of three parts. GAN loss for reconstructed samples and generated samples, recognition loss for reconstructed samples and generated samples, and the reconstruction loss.
\begin{align}
	L_{\mathrm{gen}} &= \mathsf{E} _{z \sim q(z|x) , c \sim p(c|x)}[-\log  D(G(z,c))  \nonumber \\ & 
	\qquad \qquad \qquad \qquad \qquad - \lambda_{2} \log  Q(c|G(z,c)] \nonumber \\ & \: \: \:
	           +E _{z \sim p(z) , c \sim p(c)}[-\:\log \: D(G(z,c))  \nonumber \\ &  
	           \qquad \qquad \qquad \qquad \qquad  -\lambda_{2} \log  Q(c|G(z,c) ] \nonumber \\ &  \: \: \:  
	           +\lambda_{1} L_{\mathrm{recon}} ,
\end{align}
where $\lambda_{1}$ and $\lambda_{2}$ are the hyper parameters, that quantify the relative importance between reconstruction and recognition objectives. 
    
    Finally, the discriminator is trained with three sets of examples with images from the actual data set treated as real examples and generated and reconstructed images treated as fake examples. Discriminator loss, $L_{\mathrm{dis}}$ is 
\begin{align}
	L_{\mathrm{dis}} &= \mathsf{E} _{z \sim q(z|x) , c \sim p(c|x)}[\:\log \: D(G(z,c))] \nonumber \\ & \: \: \: + 
               \mathsf{E} _{z \sim p(z) , c \sim p(c)}[\:\log \: D(G(z,c))] \nonumber \\ & \: \: \: \: \: \mathsf{E}_{x\sim p(x)}[-\:\log \: D(x)]. 
\end{align}
Given these losses, steps for training the ViGAN is as follows: 
\begin{enumerate}\itemsep0.5pt
\item minimize $L_{\mathrm{enc}}$ w.r.t to parameters of encoder,
\item minimize $L_{\mathrm{gen}}$ w.r.t to parameters of generator, 
\item minimize $L_{\mathrm{recog}}$ w.r.t to parameters of recognizer, and
\item minimize $L_{\mathrm{dis}}$ w.r.t to parameters of discriminator. 
\end{enumerate}
This setup with VAE and GAN sharing common decoder/generator is similar to the model described in \cite{AEBP}. However, the decoder/generator  conditioned on encoded representation of the image and the attribute vector. Further, to improve the perceptual quality of images, we consider the reconstruction loss in the hidden space of the discriminator instead of the pixel space as in \cite{AEBP}.  

One of the main advantage of ViGAN is that the same network can be used to generate new images conditioned on attributes as well modify a given image by fixing the image representation and changing attributes. The procedure for this is described below. 

\begin{itemize}
    \item To generate new images, sample $ z $ from $p(z)$and $c$ from $P(c|x)$. Then pass the concatenated vector $[z,c]$ through generator. 
    \item To modify a image, first get representation of image $z$ by  by passing it through encoder. Then sample the desired attributes $c$. Then pass the concatenated vector $[z,c]$ through generator. 
\end{itemize}

\section{Experiments}
\subsection{Datasets}
        We trained the proposed model on following three datasets.
        
    \begin{enumerate}
        \item MNIST: We created a modified version of MNIST by placing two digits on 64 x 64 grid. First digit is placed randomly in left half of the grid and second digit in right half of the grid.  
        \item CelebA : This dataset consists of around 200,000 images of human faces. Each images is paired with 40 binary attributes such as male, smiling, eyeglasses, mustache etc. We resized each image to 64 x 64 pixels.
        \item LFW: This dataset consists of around 13,000 images of human faces paired 73 real valued visual attributed. We cropped and scaled each image to 64 x 64 pixels.
       
    \end{enumerate}

\subsection{Training Details}
    We used a similar architecture for all three datasets, with appropriated changes to the dimension of the attribute vector and number of image channels (one channel for modified MNIST and three for other two datasets). 
    Detailed architectures for encoder, generator and discriminator are presented in the following tables
\begin{center}
\begin{tabular}{ | P{8cm}| }
	\hline
	Encoder \\  \hline
	Input: 64 x 64 x 3 or 64 x64 x 1 image \\ \hline
	4 x 4 x 64 conv, stride 2 , leaky relu \\ \hline
	4 x 4 x 128 conv, stride 2 , leaky relu, batchnorm \\ \hline
	4 x 4 x 256 conv, stride 2 , leaky relu, batchnorm \\ \hline
	512 fully connected, leaky relu, batchnorm \\ \hline
	512 fully connected, tanh, batchnorm \\ \hline
\end{tabular}
\end{center}

\medskip

\begin{center}
\begin{tabular}{ | P{8cm}| }
	\hline
	Generator /  Decoder \\  \hline
	Input: 256 + 40 for celebA, 256 + 73 for LFW , \\ 256 + 20 for MNIST \\ \hline
	4 x 4 x 448 fully connected, relu, batchnorm \\ \hline
	4 x 4 x 256 deconv, stride 2 , relu, batchnorm \\ \hline
	4 x 4 x 128 deconv, stride 2 , relu \\ \hline
	4 x 4 x 64 deconv, stride 2 , relu \\ \hline
	4 x 4 x 3(1) deconv, stride 2 , tanh \\ \hline
\end{tabular}
\end{center}

\begin{center}
\begin{tabular}{ | P{8cm}| }
	\hline
	Discriminator \ Recognizer \\  \hline
	Input: 64 x 64 x 3 or 64 x 64 x 1 image \\ \hline
	4 x 4 x 64 conv, stride 2 , leaky relu \\ \hline
	4 x 4 x 128 conv, stride 2 , leaky relu, batchnorm \\ \hline
	4 x 4 x 256 conv, stride 2 , leaky relu, batchnorm \\ \hline
	Discriminator: 1 fully-connected  \\
	Recognizer: 128 fully connected, leaky relu, batchnorm  \\
				40 fully connected, sigmoid for celebA \\
				73 fully connected, sigmoid for LFW \\ 
				20 fully connected, sigmoid for MNIST \\ \hline
	
\end{tabular}
\end{center}
\vfill 
    For optimization, Adam optimizer \cite{adam} is used for all the networks, with learning rates of  0.001, 0.001, 0.0002, 0.0002 for encoder, generator, discriminator and recognizer respectively. 
    
    We observed the ability of the network to modify images, to be very sensitive to the choice of hyper parameters $ \lambda_{1} $ and $ \lambda_{2} $. Suitable choice of $ \lambda_{1} $ and $ \lambda_{2} $ is very important to strike a balance between ability of network to accurately reconstruct the input and at the same time be sensitive to changes in the attributes. 
    
\subsection{Results}
Below we demontrate the ability of network to generate new images, as well as to modify a given image by changing the attributes. Figure 2, shows the samples from MNIST dataset and their corresponding reconstructions. 
\vspace{0.5cm}

\begin{figure}[htb]
\begin{minipage}[b]{1\linewidth}
  \centering
  \centerline{\includegraphics[width=8.0cm]{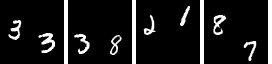}}
  \centerline{(a) Samples from modified MNIST }
\end{minipage}

\vspace{.5cm}
\begin{minipage}[b]{1\linewidth}
  \centering
  \centerline{\includegraphics[width=8.0cm]{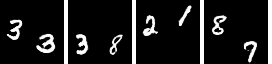}}
  \centerline{(b) Reconstructions on modified MNIST}
\end{minipage}

\caption{Reconstruction results on MNIST dataset }
\label{fig:fig1}
\end{figure}

In Figure 3 and Figure 4, we demonstrate the capability of the proposed model to modify images according to the changes in attributes, while keeping other features intact. In Figure 3. the middle row shows the actual images, while in top and bottom rows, first digit is replaced by 0 and 1 respectively. Similarly in Figure 4, the middle row shows the actual images, while in top and bottom rows, second digit is replaced by 2 and 7 respectively.

\begin{figure}[htb]
\begin{minipage}[b]{1\linewidth}
  \centering
  \centerline{\includegraphics[width=8.0cm]{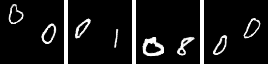}}
  \centerline{(a) First digit replaced by 0}
\end{minipage}

\vspace{.5cm}
\begin{minipage}[b]{1\linewidth}
  \centering
  \centerline{\includegraphics[width=8.0cm]{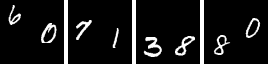}}
  \centerline{(b) Samples from modified MNIST dataset}
\end{minipage}
\vspace{.5cm}

\begin{minipage}[c]{1\linewidth}
  \centering
  \centerline{\includegraphics[width=8.0cm]{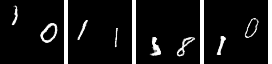}}
  \centerline{(c) First digit replaced by 1}
\end{minipage}
\vspace{.5cm}

\caption{Modification results on modified MNIST dataset with first digit replaced by 1 and 0 in top and bottom rows respectively, while preserving the location of both the digits. Middle row shows the actual images. }
\label{fig:fig2}
\end{figure}

\begin{figure}[htb]
\begin{minipage}[b]{1\linewidth}
  \centering
  \centerline{\includegraphics[width=8.0cm]{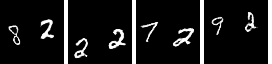}}
  \centerline{(a) Second digit replaced by 2}
\end{minipage}

\vspace{.5cm}
\begin{minipage}[b]{1\linewidth}
  \centering
  \centerline{\includegraphics[width=8.0cm]{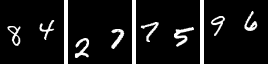}}
  \centerline{(b) Samples from modified MNIST dataset}
\end{minipage}
\vspace{.5cm}

\begin{minipage}[c]{1\linewidth}
  \centering
  \centerline{\includegraphics[width=8.0cm]{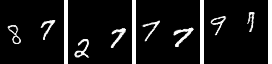}}
  \centerline{(c) Second digit replaced by 7}
\end{minipage}
\vspace{.5cm}

\caption{Modification results on modified MNIST dataset with second digit replaced by 2 and 7 in top and bottom rows respectively, while preserving the location of both the digits. Middle row shows the actual images.}
\label{fig:fig3}
\end{figure}

\begin{figure}[htb]
\begin{minipage}[b]{1\linewidth}
  \centering
  \centerline{\includegraphics[width=8.0cm]{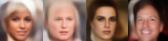}}
  \centerline{(a) Samples from VAE }
\end{minipage}

\vspace{.5cm}
\begin{minipage}[b]{1\linewidth}
  \centering
  \centerline{\includegraphics[width=8.0cm]{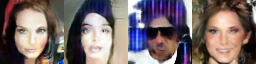}}
  \centerline{(b) Samples from GAN}
\end{minipage}
\vspace{.5cm}
\begin{minipage}[b]{1\linewidth}
  \centering
  \vspace{0.5cm}
  \centerline{\includegraphics[width=8.0cm]{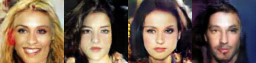}}
  \centerline{(c) Samples from the proposed model}
\end{minipage}
\caption{Qualitative comparison between VAE, GAN and the proposed model. Samples from proposed model appear more realistic than the samples from VAE and GAN }
\label{fig:fig4}
\end{figure}

In Figure 6, eyeglasses are added to the base image from celebA dataset by changing the corresponding attribute value to 1. Figure 7. shows the modification results on LFW dataset with real valued attributes. In Figure 4. we present four variations of two base images, generated by the network, corresponding to the attributes frowning, smiling with closed mouth, smiling with open mouth and wide eyes.

\begin{figure}[htb]
\begin{minipage}[b]{1\linewidth}
  \centering
  \centerline{\includegraphics[width=8.0cm]{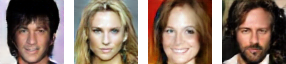}}
  \centerline{(a) Reconstructions on celebA dataset}
\end{minipage}
\vspace{.5cm}
\begin{minipage}[b]{1\linewidth}
  \centering
  \vspace{0.5cm}
  \centerline{\includegraphics[width=8.0cm]{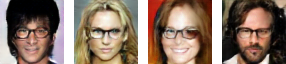}}
  \centerline{(c) Reconstructions with Eyeglasses bit turned on}
\end{minipage}
\caption{Modification results on celebA dataset }
\label{fig:fig4}
\end{figure}

\begin{figure}[htb]
\begin{minipage}[b]{1\linewidth}
  \centering
  \centerline{\includegraphics[width=8.0cm]{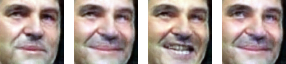}}
\end{minipage}

\begin{minipage}[b]{1\linewidth}
  \centering
  \vspace{0.5cm}
  \centerline{\includegraphics[width=8.0cm]{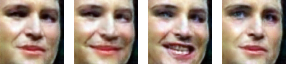}}
\end{minipage}
\caption{Modification results on LFW dataset. From left to right, frowning, smiling, mouth open, wide eyes }
\label{fig:fig5}
\end{figure}

\section{Concluding Remarks}

Several deep generative models have been proposed in the recent years. Some of the successful approaches are Variational Auto Encoders (VAEs), Generative Adversarial Networks (GANs), Deep Recurrent Attention Writer (DRAW) , Pixel Recurrent Neural
Networks (PixelRNN). All these frameworks are usually trained with back propagation.  

Further some extensions of above mentioned models are proposed to generate images from text. Some examples of such models are \cite{attention, cGAN, cGANd} . These models can generate images given attributes, but cannot modify a given image. 

\cite{AEBP} proposed a method to modify faces, by adding or subtracting a mean visual attribute vector. But the method is applicable for only binary attributes. \cite{gaurav} proposes another method based on embedding visual representations and attribute descriptions in a common space.

Recently Reed et.al \cite{cGAN} also proposed a similar method based for conditional image generation. They too demonstrate image modification with their network. But they learn the posterior separately from the generation process, while our method learns posterior along with the generation network 

In the work, we proposed a new architecture for joint modeling of images and their
visual descriptions by combining the ideas from Variational Auto Encoders (VAEs) and InfoGAN. We demonstrate the ability of the proposed model to generate new images as well as to modify existing images.

\newpage
\newpage
\bibliography{paper}
\bibliographystyle{icml2016}


\end{document}